\documentclass[10pt,twocolumn,letterpaper]{article}

\usepackage[pagenumbers]{cvpr} %
\usepackage{float}
\usepackage{booktabs}
\usepackage{wrapfig}
\usepackage{listings}
\usepackage{algorithm}
\usepackage{algpseudocode}
\usepackage{amsmath}
\usepackage{svg}
\usepackage{graphicx}
\usepackage{subcaption}
\usepackage{multirow}

\makeatletter
\@namedef{tabular*}#1{\footnotesize%
 \setlength\dimen@{#1}%
   \edef\@halignto{to\the\dimen@}\@tabular}

\def\toprule{\\[-6.5pt]\Hline\\[-6.45pt]}
\def\colrule{\\[-6.5pt]\Hline\\[-6.55pt]}
\def\botrule{\\[-6.6pt]\Hline\\[-5.2pt]}
\def\Hline{%
  \noalign{\ifnum0=`}\fi\hrule \@height .43pt \futurelet%
   \@tempa\@xhline}
\makeatother

\definecolor{cvprblue}{rgb}{0.21,0.49,0.74}
\usepackage[pagebackref,breaklinks,colorlinks,citecolor=cvprblue]{hyperref}

\title{Industrial Language-Image Dataset (ILID):~\\Adapting Vision Foundation Models for Industrial Settings}

\author{
Keno Moenck\textsuperscript{\rm 1,}\thanks{Corresponding author: \href{mailto:keno.moenck@tuhh.de}{keno.moenck@tuhh.de}} \quad Duc Trung Thieu\textsuperscript{\rm 1} \quad Julian Koch\textsuperscript{\rm 1} \quad Thorsten Sch{\"u}ppstuhl\textsuperscript{\rm 1} \vspace{5pt}\\
{\normalsize \textsuperscript{\rm 1}{Hamburg University of Technology, Institute of Aircraft Production Technology}} \vspace{5pt}\\
\small{\href{https://github.com/kenomo/ilid}{github.com/kenomo/ilid}}
}

\begin{document}
\maketitle

\begin{abstract}
In recent years, the upstream of Large Language Models (LLM) has also encouraged the computer vision community to work on substantial multimodal datasets and train models on a scale in a self-/semi-supervised manner, resulting in Vision Foundation Models (VFM), as, e.g., Contrastive Language–Image Pre-training (CLIP). The models generalize well and perform outstandingly on everyday objects or scenes, even on downstream tasks, tasks the model has not been trained on, while the application in specialized domains, as in an industrial context, is still an open research question. Here, fine-tuning the models or transfer learning on domain-specific data is unavoidable when objecting to adequate performance.
In this work, we, on the one hand, introduce a pipeline to generate the Industrial Language-Image Dataset (ILID) based on web-crawled data; on the other hand, we demonstrate effective self-supervised transfer learning and discussing downstream tasks after training on the cheaply acquired ILID, which does not necessitate human labeling or intervention. With the proposed approach, we contribute by transferring approaches from state-of-the-art research around foundation models, transfer learning strategies, and applications to the industrial domain.
\end{abstract}

\section{Introduction}

Machine vision technologies facilitated by deep learning usually outperform traditional methods, especially in dynamic and open settings. In the scope of training deep models, industrial contexts\footnote{We define the industrial domain as follows: industrial activities serve to produce consumable goods or a capital asset, which includes production as the superordinate term involving all processes around it, including activities from manufacturing, assembly, logistics, or finance. In addition, tasks in the later lifecycle of a product, like Maintenance, Repair, and Overhaul (MRO), also belong to industrial activities. Vision applications are typically closer to the shopfloor than the topfloor.} lack everyday objects and scenes, typically covered by publicly available datasets, which is why applications in these specialized domains here demand custom datasets, e.g., synthetically generated \cite{Schoepflin.2021b,Schoepflin.2022b,Holst.2022,Schmedemann.2022}, which model the specific object and sensor domain. 

The availability of curated, publicly accessible datasets specific to industrial needs is exceedingly sparse, e.g., the MVTec \cite{Drost.2017,Bergmann.2019,Bergmann.2021,Bergmann.2022}, VISION \cite{Bai.13.06.2023}, or tool recognition \cite{Busch.2023} datasets encapsulate only a limited spectrum of objects and support only a handful of trainable tasks based on the provided ground truth annotations.
Besides the need for training data, fine-tuning, domain adaptation, or transfer learning, transferring a model from a source to a related target, e.g., object/scene/sensor domain, is ineluctable, which can reduce the necessary samples per conceptual class to only a few shots during training. The model's pre-training is the critical point here, where training data size, variability, and model size directly relate to the overall performance \cite{Zhang.2024}.

Large-scale pre-trained foundation models represent a paradigm shift in Artificial Intelligence (AI), characterized by extensive self-supervised training \cite{Bommasani.16.08.2021}. These models, e.g., BERT \cite{Devlin.11.10.2018}, the well-known GPT-n series \cite{ Budzianowski.12.07.2019,Brown.28.05.2020,OpenAI.15.03.2023}, or Llama \cite{ Touvron.27.02.2023,Touvron.18.07.2023,MetaAI.26.05.2024}, learn rich knowledge representations capable of transcending to various downstream tasks. 
The shift in AI drives single tasks and single-modalities learners to a paradigm encompassing diverse tasks and multimodalities, which more closely mimics human perception and cognitive processes. Following Large Language Models (LLM), Vision Foundation Models (VFM) have been upstreamed in the last few years, capable of supporting not only 2D or even 3D modalities but also language \cite{Awais.2023}.
Data for training at scale is typically web-crawled from the vast resources of the Internet, which then demands sophisticated post-processing, posing a variety of challenges \cite{Schuhmann.16.10.2022,Changpinyo.17.02.2021}.
Besides, given such large, partially unstructured datasets, only self-supervised or unsupervised methods are able to learn from the data effectively.

A self-supervised approach capable of learning from text and image modalities is contrastive learning, in which a model learns to distinguish between positive and negative combinations of samples, firstly, nearly concurrently, presented by CLIP \cite{Radford.2021} and ALIGN \cite{Jia.2021} at a large scale.
Contrastive learning by contrasting positive and negative samples in a batch, in the case of vision and language, is based on a text and image encoder. The idea is that the encoders are trained to output embeddings for the image and text, increasing the similarities of positive samples by decreasing the distance in the joint embedding space and increasing the distance of negative samples.
Employing a text encoder allows for natural language supervision, relaxing the necessity of fixed classes as in the case of training traditional deep learning models like a ResNet \cite{He.10.12.2015}. This fact makes assembling a dataset at a scale less laborious since assigning an image to a fixed class omits, enabling learning from unstructured data.
Different language-image datasets of scale have emerged, ranging from 12M \cite{Changpinyo.17.02.2021} to 5B \cite{Schuhmann.16.10.2022} samples. Since they are based on web-available data, not all cleaned, post-processed, and curated datasets are published, as in the case of CLIP.

\begin{figure}[!h]
    \centering
    \includegraphics[width=1.0\linewidth]{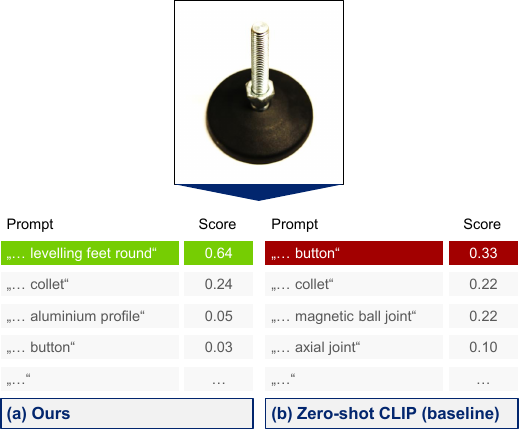}
    \caption{CLIP on the task of classification after (a) transfer learning on the Industrial Language-Image Dataset (ILID) and (b) the zero-shot baseline results.}
    \label{fig:before_after}
\end{figure}

VFMs exhibit rich knowledge representations, are adaptable to various downstream tasks, and generalize better than conventional models, but only to a certain extent in novel and out-of-distribution domains, necessitating fine-tuning or transfer learning. As demonstrated in Fig. \ref{fig:before_after}, the zero-shot model CLIP, given a highly out-of-distribution image, does not predict nor even close to the ground truth.
As already outlined, in the industrial domain, we face non-everyday objects and scenes, which is why we can not rely on commonly available datasets for fine-tuning or transfer learning, which also inhibits the use of VFM here. In this work, we try to make a step in the direction of utilizing VFM capabilities in specialized industrial domains by contributing three-folded:
\begin{itemize}
    \item We propose a method to generate the Industrial Language-Image Dataset (ILID) from web-crawled data and release a version that covers objects from different industrial-related domains\footnote{Since the data from the web do not belong to us, we are not allowed to publish the images and texts, but we provide the final post-processed metadata, which can be used to reassemble the dataset. Please contact the corresponding author.}. We publish the pipeline to generate the ILID at \href{https://github.com/kenomo/ilid}{github.com/kenomo/ilid}.
    \item We effectively demonstrate transfer learning to CLIP with the given dataset, which outperforms CLIP’s zero-shot capabilities.
    \item We elaborate on different tasks that serve industrial domain-related vision applications. We publish the training- and evaluation-related code here \href{https://github.com/kenomo/industrial-clip}{github.com/kenomo/industrial-clip}.
\end{itemize}

This work focuses on utilizing CLIP rather than other vision-language models due to the significant established usage and fine-tuning/transfer learning strategies. Besides, comparing only one established model on the data increases the focus, clarity, and depth of the findings in the scope of this work. Nevertheless, we encourage the reuse of ILID with other strategies or also employ further fine-tuning and transfer learning strategies.
~\\

The rest of this work is structured as follows: First, we outline in Sec. \ref{sec:related_works} existing applications of VFMs in industrial applications, introduce Contrastive Image-Language Pre-training (CLIP) and current existing fine-tuning/transfer-learning approaches. In Sec. \ref{sec:method}, we present our overall method of generating the dataset as well as our training procedure. We elaborate on our extensive experiments in Sec. \ref{sec:experiments}. We conclude and discuss this work in Sec. \ref{sec:conclusion_outlook}.

\begin{figure*}[!ht]
    \centering
    \includegraphics[width=\linewidth]{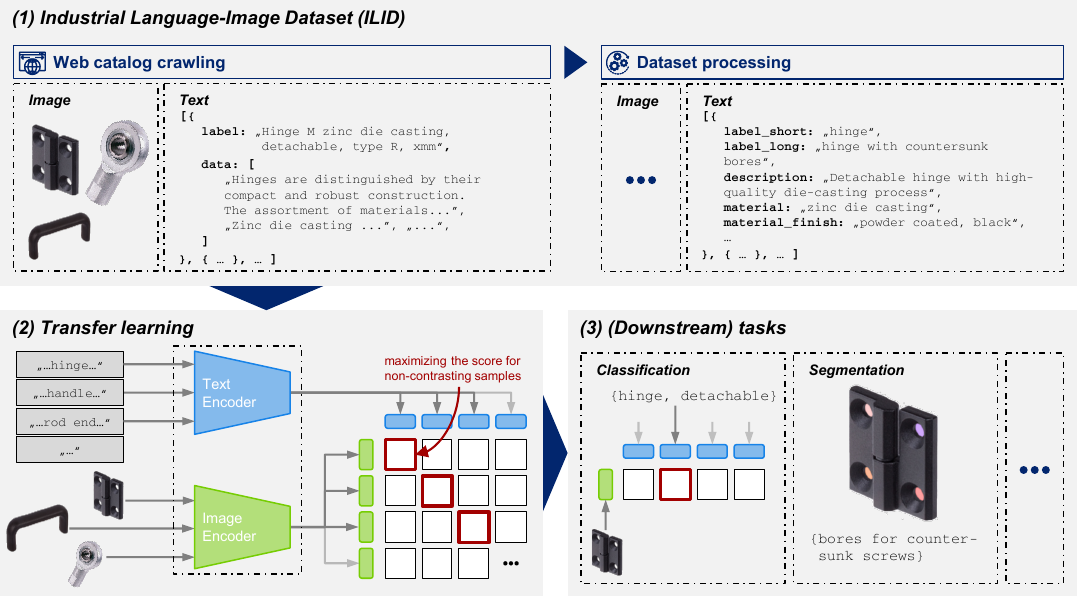}
    \caption{Overview of this work's method: (1) generation of the Industrial Language-Image Dataset (ILID), (2) transfer learning using the ILID, and (3) evaluating the performance in different tasks.}
    \label{fig:overview}
\end{figure*}

\section{Related Works}\label{sec:related_works}

\subsection{VFMs in industrial applications}\label{sec:vfm_industrial_applications}

Code recognition, object or position recognition, completeness, shape/dimension check, or quantitive or qualitative inspection are typical vision applications in manufacturing \cite{Hornberg.2017}. While in manufacturing, these are often suited toward narrow fields of view and close to the object; in the neighboring domain, intralogistics, tasks are besides close ones, like inspecting load carriers for trash, contamination, and damage or documentation, verification, assistance, and automation, perceiving the environment is often of interest, which results in, e.g., foreign debris detection or tracking objects \cite{Naumann.12.04.2023,Moenck.2023}. 
The first step in the perception pipeline of these applications is typically a fundamental vision task, e.g., in the 2D domain, giving each pixel semantic and clustering pixel to semantically meaningful regions. Then follows additional enhancing the output with further semantics and finally forming the application-specific decision used in, e.g., part of a production system.

Up to this date, there exists only a small set of publications on the topic of utilizing VFMs in one or more steps of such vision pipelines, which we give a small excerpt in the following: On a broader scale \cite{Wang.2023b} explore use cases for deploying VFMs in the industrial context without designing or elaborating on specific architectures and how to train, fine-tune, or do transfer learning. \cite{Moenck.2023} discusses the abilities of the Segment Anything Model (SAM), a class-agnostic segmentation model that outstandingly generalizes to unseen objects and scenes, in the context of vision applications in the aircraft industry, including manufacturing, intralogistics, and MRO. \cite{Zhang.2023} name two use cases in PCB defect inspection and industrial human action recognition.

Current literature throws up ideas on utilizing LLMs, e.g. \cite{Makatura.25.07.2023}, or VFMs, e.g., \cite{Moenck.2023,Wang.2023b,Zhang.2023,Picard.21.11.2023}, in the industrial domain; little is known about how to enable VFM to perform effectively in specific use cases. Besides having suitable datasets, training with the data demands specific strategies. We will elaborate on the aspects in the following sections.

\subsection{Contrastive Language-Image Pre-training (CLIP)}\label{sec:clip}
CLIP learns rich image-text representations from natural language supervision utilizing natural language as a prediction space to reach higher performance in generalization and transfer. It is not an entirely novel approach; however, the origin of the idea of learning from perceptions in natural language is not exactly dated to specific research.
In 1999, \cite{Mori.1999} explored retrieving words for unknown images based on statistical learning to predict nouns and adjectives. In 2007, \cite{Quattoni.2007} demonstrated learning image representations using manifold learning to predict words for images. Recent approaches that emerged before CLIP and learn visual representations from text are Visual representations from Textual annotations (VirTex) \cite{Desai.11.06.2020}, Image-Conditioned Masked Language Modeling (ICMLM) \cite{Sariyildiz.04.08.2020}, and Contrastive Visual Representation Learning from Text (ConVIRT) \cite{Zhang.02.10.2020}.

\subsubsection{Contrastive learning}
\begin{figure}[!h]
    \centering
    \includegraphics[width=0.8\columnwidth]{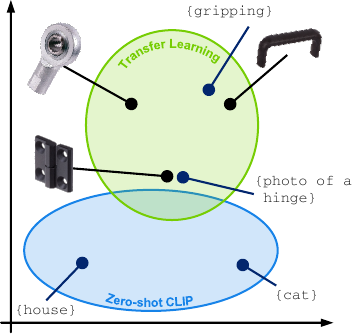}
    \caption{Joint embedding space of text and image representations: conceptually similar texts and images are encoded close to each other, dissimilar pairings do not share similar positions.}
    \label{fig:idea_embedding_space}
\end{figure}

A contrastive learning model consists of two main components: (1) an encoder for all input modalities and (2) a loss function measuring the similarity between positive and negative pairs. The encoder can be reused from other models and training, e.g., demonstrated by OpenScene \cite{Peng.2022}, which employs a frozen text and 2D image encoder while training a 3D point cloud encoder for language-based 2D/3D scene understanding.
The encoder models are trained to complement and comprehend each other fully by encoding similar concepts of images and text in similar vectors. That is, a text representing "photo of a hinge" would output a similar vector as the image counterpart and be further away from images that are not connected, as shown in Fig. \ref{fig:idea_embedding_space}. Besides prompting for the object's name, a sufficiently trained text encoder would encode, e.g., conceptual close activities near the object's name embedding (s. Fig. \ref{fig:idea_embedding_space}).

The self-supervised pre-training of CLIP followed: Given \(N\) pairs of image and text, CLIP estimates the similarity for all the possible \(N \times N\) pairings. With each text and image pair in the multimodal vector space, the models inside CLIP are jointly trained to maximize the similarity of each positive pairing and, at the same time, minimize the similarity of \(N \times N - N\) antagonistic pairs (s. Fig. \ref{fig:overview}). The embedding similarities between pairs are represented by the cosine similarity metric, which is used to optimize the cross-entropy loss in order to build the most optimized versions of both the image and text encoder at the same time.

\subsubsection{Performance}\label{sec:clip_performance}
Zero-shot CLIP achieves similar performance or even outperforms conventional fully supervised class-wise models while preserving robustness through the ability to learn from a broader range of representations from natural language, especially on in-distribution or slightly out-of-distribution data. On the other hand, zero-shot CLIP weakly performs on datasets that are far out-of-distribution, such as satellite images (EuroSAT \cite{Helber.31.08.2017}, NWPU-RESISC45 \cite{Cheng.2017}) or tumors (PatchCamelyon \cite{Veeling.08.06.2018}) \cite{Radford.2021}. 
When comparing CLIP and other large pre-trained n-shot models such as BiT-M \cite{Kolesnikov.24.12.2019} and SimCLRv2 \cite{Chen.17.06.2020}, CLIP's authors depict that the zero-shot performance outperforms all other models on the metric of average accuracy score up to 4-shot linear classifiers trained on the same dataset \cite{Radford.2021}.
The limitations are that scaling the model to learn from more data has steadily increased the performance, but computing power increases exponentially, which is currently barely economically reasonable.

\subsubsection{Recent development}
Meanwhile, much work exists on further development and adaptations of CLIP \cite{Li.11.10.2021,Goel.28.05.2022,Hu.04.08.2023,Yu.04.05.2022,Rao.02.12.2021,Yao.09.11.2021,Mu.23.12.2021,Sun.27.03.2023}.
The most notable works are SLIP \cite{Mu.23.12.2021}, DeCLIP \cite{Li.11.10.2021}, ReCLIP \cite{Hu.04.08.2023}, CoCa \cite{Yu.04.05.2022}, and FILIP \cite{Yao.09.11.2021}, aiming to improve efficiency in the training process. 
SLIP combines language supervision and image self-supervision to improve performance further. DeCLIP employs supervision across modalities as well as self-supervision within each modality, whereas ReCLIP at first learns pseudo labels and then applies cross-modality self-supervision. 
CoCA, on the other hand, skips cross-attention in some decoder layers to encode unimodal text representations and cross-attend the remaining layers with the image encoder. By using contrastive loss between unimodal image and text embeddings, along with captioning loss for multimodal decoder outputs, CoCa efficiently trains on a wider variety of data with minimal overhead. 
Improved fine-grained performance of CLIP is demonstrated in the works of FILIP \cite{Yao.09.11.2021}, where instead of contrastive loss being calculated from global features of an entire image and text sequence, token-wise cross-modal interaction is modeled to take into account image patches and textual tokens more fine-grained.

Since this work focuses mainly on the training data, we will not evaluate all the individual strategies that aim to increase performance. Instead, we use the vanilla CLIP model and employ basic transfer learning methods that we can employ with limited hardware resources, which also demonstrate the effectiveness in the scope of lower-cost applications.

\subsection{Fine-tuning and transfer learning}\label{sec:fine-tuning_transfer-learning}
Depending on the application, CLIP has two different ways to adapt to a new distribution, i.e., new sets of data entirely outside the dataset on which CLIP was pre-trained.
Fine-tuning and transfer learning are very similar ways to adapt CLIP, but they have different applications depending on the task at hand and different processes in modifying the architecture. Fine-tuning consists of training all layers or at least parts of the model. This process is usually more suitable for adapting to small sets of data that are closely related to the dataset CLIP was pre-trained on, such as everyday objects and general concepts.
On the other hand, in tasks where the dataset is too specific, i.e., specialized knowledge, transfer-learning is better suited, as it freezes all the original layers of the pre-trained model and only adds or injects extra trainable layers or parameters. This way, the learned features of the zero-shot model are preserved and optimized for generalization to novel, previous out-of-distribution data. 
Usually, the fine-tuning process requires much more resources in terms of time, data, and computation as it modifies all the layers of the model compared to transfer learning. In the case of ILID, transfer learning proved to be a fitting solution, as the dataset is specialized specifically on industrial components, which are not presumably contained in CLIP's dataset used for pre-training.

Notable works in transfer-learning of CLIP are adapter-styled tuning, e.g., CLIPAdapter \cite{Gao.2021}, 
 and prompt learning, e.g., CoOp \cite{Zhou.2021} and APEX \cite{Yang.11.27.2023}.
CLIPAdapter (s. Fig. \ref{fig:model_architectures}) adds dense down- and up-sampling layers on top of CLIP either to the image, text, or both encoders. Thereby, only the most prominent features are compressed into lower dimensions. From the latent space, the adapter then learns to reconstruct the essential ones.
CoOp (s. Fig. \ref{fig:model_architectures}) is the first to demonstrate continuous prompt learning as introduced by \cite{Lester.18.04.2021} for CLIP, which is learning continuous prompts by backpropagation for each label or one specific prompt template for all labels. Concretely, CoOp creates a set of learnable vectors, initialized by random values or given text embeddings, which, during training, the model adapts to.
APEX is the most recent approach that also evaluates adding learnable tokens to the residual transformer blocks in the image encoder. Besides, APEX introduces a residual connection skipping the text adapter steered by an adaptive coefficient to perform better on a variety of out-of-distribution data.

\section{Method}\label{sec:method}

In this section, we first outline the generation of the ILID, including a thorough outline of the dataset acquisition, the criteria for data selection, web crawling to gather extensive sets of unlabeled data, and filtering (s. Sec. \ref{sec:pipeline}). Secondly, we elaborate on the decision for the model architecture and training procedure in Sec. \ref{sec:transfer_learning}.

\subsection{Dataset generation pipeline}\label{sec:pipeline}

\begin{figure}
    \centering
    \includegraphics[width=1.0\columnwidth]{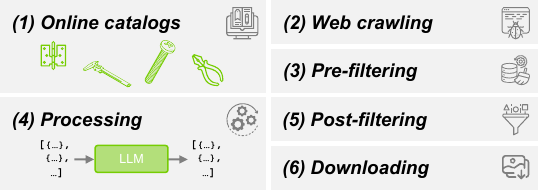}
    \caption{Dataset generation pipeline resulting in the Industrial Language-Image Dataset (ILID).}
    \label{fig:dataset_pipeline}
\end{figure}

Following a typical data pipeline structure, including data selection, transforming, and pre-/post-filtering (s., e.g., \cite{Changpinyo.17.02.2021}), we employed six steps (s. Fig. \ref{fig:dataset_pipeline}) to generate the Industrial Language-Image Dataset (ILID). Each of the steps results in a structured JSON document containing all the outputs. The next step always takes the respective document as input.

\begin{enumerate}
    \item While searching for reasonably organized industrial-related data on the Internet, we found that \textbf{online catalogs} contain relevant language-image information. Typically, web stores have a page per product, sometimes imaging a set of product configurations, a precise, often standardized, title, description, information about the material, and further information about the product. These online stores are an adequate data source for the industrial domain. The first step was identifying a store set containing the necessary object-domain.
    
    \item \textbf{Web crawling} data from online catalogs follows two basic steps: getting the sitemap from \textit{robots.txt} and writing a crawler for the specific structure of the product pages. The top-level \textit{robots.txt} file delineates the \textit{Robots Exclusion Protocol}, which guides crawlers and other bots on which sections of the website they are permitted to access. Typically, this file also specifies the location of the sitemap, an XML-formatted document designed to provide crawlers with information about all pages on a website. Sitemaps can be hierarchically ordered; in the case of online catalogs, typically, there is one specific sitemap containing all products and their respective locations. We use Scrapy\footnote{\href{https://scrapy.org/}{Scrapy: A Fast and Powerful Scraping and Web Crawling Framework}} as a Python-based web crawler that takes a sitemap as input and crawls through all the specified locations. Creating a specific spider for a web catalog requires manual intervention since one has to define which images and text blocks to yield. Besides a central label tag for each entry, we save an unstructured list-typed data object, which can contain all other available information about the product, like materials, finish, colors, etc.
    Using the sitemap as the initial crawling entry point is a common step in every online search engine.
    
    \item In the \textbf{pre-filtering} step, we filter for duplicate entries, remove special characters, as well as diminish entries that do not have sufficient information. Besides, we filter the data for a set of trade names and remove these from all product information. Often, industrial product names include the manufacturer, which we do not want to use further or bias the data within the following information extraction.
    
    \item In the central \textbf{processing} step, we use a small local-deployable LLM to extract our five target information from the unstructured data. We define these as (1) a long label describing the product, (2) a short label that is shorter than the long label, (3) a description of the product, (4) the material, (5) the finish or color of the product (s. also Fig. \ref{fig:overview}). In our study, we used Llama3-8B \cite{ MetaAI.26.05.2024} in the fine-tuned instruct version (s. \ref{app:llama-prompt} for the respective prompt). We ask the LLM not to output any numbers or sizes; additionally, we remove them from the initial data since, on the one hand, we do not expect that a 2D image task can identify or recognize any dimensional quantities given different camera positions and varying intrinsics, on the other hand, we do not want to bias the dataset with it.
    After prompting for the desired information, we extract these from the response and save them for further processing. We discard the item from the dataset if the prompt does not return sufficient output.
    
    \item In the \textbf{post-filtering} step, we again filter for any unwanted characters and do some further cleaning, like lowering words.
    
    \item In the final \textbf{downloading} step, all images are downloaded, post-processed, and resized while also assembling the final JSON specifying the dataset's text and metadata.
\end{enumerate}

With the given steps, we are able to extract a product's image and a structured set of five pieces of information. Besides, we observed that even a small model such as Llama3-8B in its instruct fine-tuned version is mostly able to extract the demanded information from the bunch of unstructured text. We show an excerpt of the dataset in \ref{app:dataset_excerpt}.

\subsection{Transfer learning}\label{sec:transfer_learning}

\begin{figure*}[!ht]
    \centering
    \includegraphics[width=0.80\linewidth]{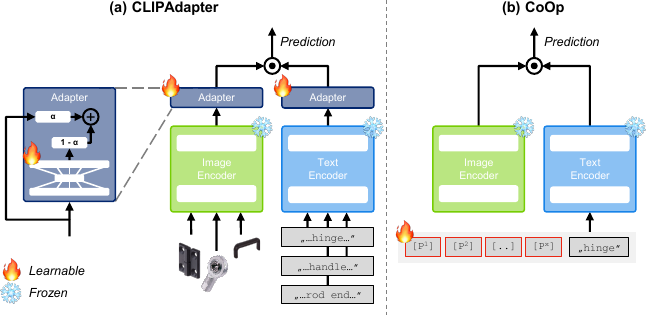}
    \caption{The architectures used in this work: (a) CLIPAdapter \cite{Gao.2021} and (b) CoOp \cite{Zhou.2021}.}
    \label{fig:model_architectures}
\end{figure*}

\subsubsection{Model architecture}

As we already outlined in Sec. \ref{sec:fine-tuning_transfer-learning}, we adopt a simple yet effective strategy for transfer learning from CLIP's in-distribution to our ILID dataset. Within CLIP's dual-encoder setup, we must utilize a strategy for the image and text stream. Fig. \ref{fig:model_architectures} depicts the used model architectures.

While we estimate that the images we want to learn but also infer from show similar characteristics as CLIP's in-distribution data compared to other fully out-of-distribution image data as in the case of, e.g., PatchCamelyon \cite{Veeling.08.06.2018} (s. Sec. \ref{sec:clip_performance}), we employ on the image stream only a simple trainable adapter as proposed by \cite{Gao.2021}. We tuned the mixing coefficient manually; we observed that a low \(\alpha\) can vastly result in overfitting, while a high value does not necessarily increase the performance significantly during cross-validation. That is why we chose a balanced value of \(\alpha = 0.5\). 
The adapters reduce the feature by \(4\) as proposed in the original paper \cite{Gao.2021}.
We omitted testing prompt tuning on the image stream as introduced by APEX \cite{Yang.11.27.2023} since we estimate a relatively low distribution shift from the CLIP dataset to ILID regarding the images.

In contrast, prompt engineering is a crucial task for learning, as well as inference with textual, promptable models. In a preliminary study, we have already observed that vanilla CLIP performs differently, given different prompt templates like \textit{"a photo of \{\}."} compared to \textit{"a photo of \{\}, an industrial product."} The difference from the minor change results from the prompts CLIP was pre-trained with, which follow similar characteristics. Having not to discretely prompt-tune manually motivated us to utilize CoOp \cite{Zhou.2021} as a continuous prompt learning method. Besides, we also evaluate in the experimental study (s. Sec. \ref{sec:experiments}) the performance of adding an additional adapter to the text stream.

\subsubsection{Training}
During the pre-training of CLIP, a very large minibatch size of \(32,768\) was used, which took for the largest Vision Transformer (ViT) configuration (428M parameters) a total of 12 days on 256 V100 GPUs \cite{Radford.2021}. Compared to the pre-training, during transfer learning with CoOp, we have a total of \(c_{n} \times 512\) trainable weights (\(c_{n} =\) number of context vectors), which can be managed on a single consumer GPU in a reasonable time. However, the batch size has to be chosen wisely from the memory point of view, as well as by looking at the dataset labels.

Given 32k samples per minibatch out of a total of 400M, the chance, utilizing random sampling, that non-contrastive samples are included in one minibatch is negligibly slight.
In contrast, fine-tuning or transfer learning approaches typically contrast all possible class labels against a set of images \cite{Gao.2021, Yang.11.27.2023, Zhou.2021, Zhou.10.03.2022} during the benchmark studies on datasets like ImageNet \cite{Deng.2009}, which is why non-contrasting samples are not possible as long as the classes are conceptually far away from each other.
The assembled ILID dataset does not have any class concept, meaning that we, as a priori, do not know how two samples and their labels are semantically close to each other. Contrasting a set of images against all possible labels is infeasible memory-wise; that is why we can not follow this training method and only contrast the images and their labels inside one batch as done during pre-training. This change led us to employ a different optimizer from the one used in the original CoOp implementation since Stochastic Gradient Descent (SGD) would not converge given the smaller batch size. We changed from vanilla SGD to Adadelta \cite{Zeiler.22.12.2012}, an SGD optimizer that adapts learning rates over time using only first-order information.

\section{Experiments}\label{sec:experiments}
In this section, we present a series of studies utilizing ILID, designed to evaluate the effectiveness of the dataset and transfer learning approach for different tasks. We begin with the dataset properties (s. Sec \ref{sec:experiments}), describe the experimental setup (s. Sec. \ref{sec:experiments_setup}), and present quantitative results on cross-validation (s. Sec. \ref{sec:experiments_quantitative_results}) as well as training and inference on a different label type (s. Sec. \ref{sec:material_prompting}).
Further, we present the results of a downstream task on segmentation (s. Sec. \ref{sec:experiments_downstream_tasks}).

\subsection{Dataset}\label{sec:experiments_dataset}
For the presented ILID, as of now, we crawled five different online shops, resulting in \(12,537\) valid samples, including a diverse range of products ranging from standard elements small in size like hinges, linear motion elements, bearings, or clamps to larger ones, like scissor lifts, pallet trucks, etc. (an excerpt is depicted in \ref{app:dataset_excerpt}).

\begin{figure}[!h]
    \centering
    \includegraphics[width=1.0\linewidth]{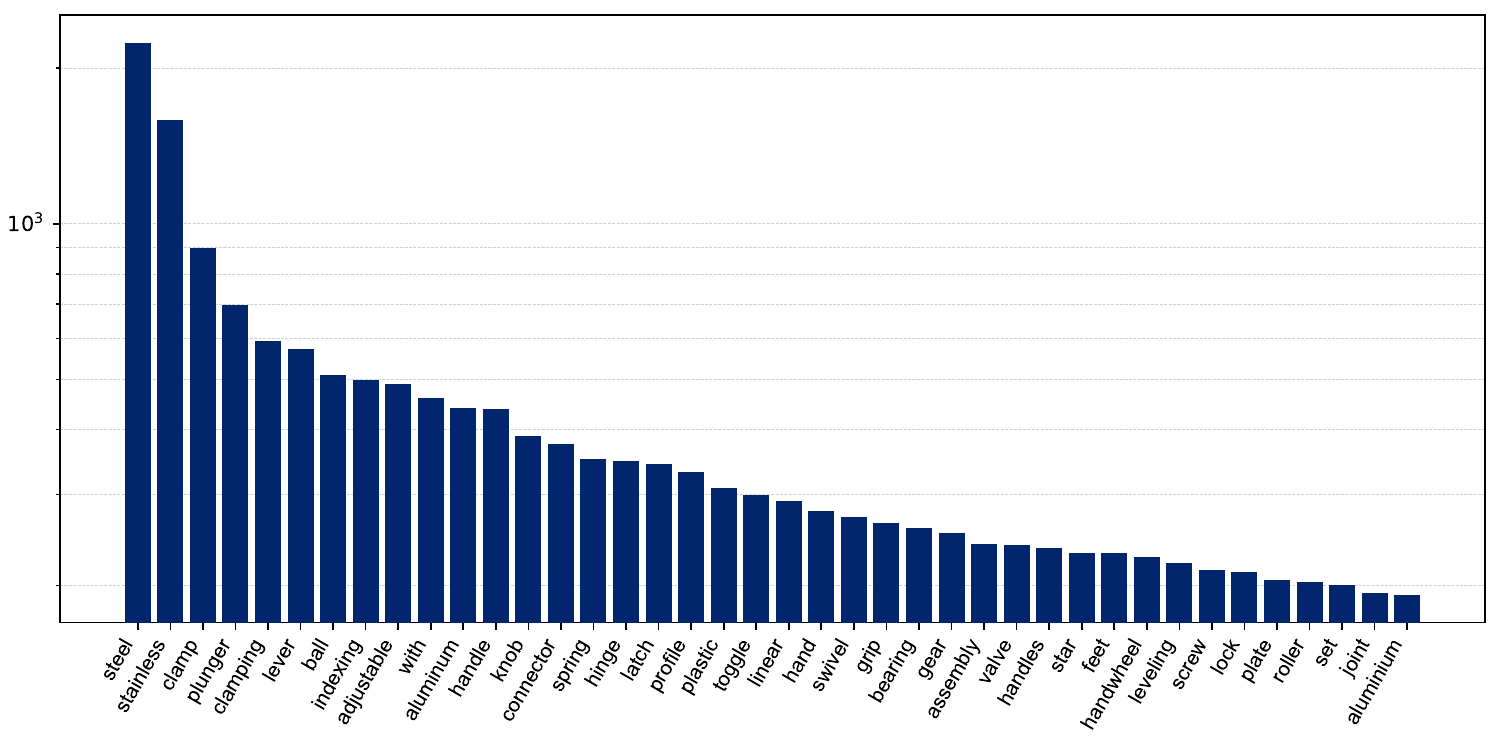}
    \caption{Top-$40$ word occurrences in label \textit{label\_short}.}
    \label{fig:word_occurrences}
\end{figure}

Fig. \ref{fig:word_occurrences} depicts the top-$40$ word occurrences in label \textit{label\_short}, showing that typical concepts of industrial standard parts like \textit{clamp}, \textit{lever}, \textit{handle}, \textit{knob}, \textit{hinge}, or \textit{swivel} are pronounced represented but also material types (\textit{steel}, \textit{aluminum} / \textit{aluminium}) and properties (\textit{stainless}) as well.

Tab. \ref{tab:dataset_characteristics} lists the number of unique labels per label category and hints at the dataset's diversity. Obviously, with increasing words (on average: \textit{label\_short} \(<\) \textit{label\_long} \(<\) \textit{description}), the number of label-wise unique labels increases. So, nearly every sample has a unique \textit{description}, but only two labels, on average, share the same \textit{label\_short}. Since we do not account for minor preposition words like \textit{a}/\textit{an}/\textit{the} in the labels, the labels are slightly more equal on the semantically level. However, we estimate a good diversity in the dataset, and since we do not account for preposition words in the counting, at least three to four samples are included per semantical similar class, which should suffice for a tuned CLIP to outperform fully supervised models (s. Sec. \ref{sec:clip_performance}).
We use the presented version of ILID in the following experiments.

\begin{table}[!h]

    \caption{Number of unique labels per label category.}
    \begin{tabular*}{\hsize}{@{\extracolsep{\fill}}ccccc@{}}
    \toprule
    \textit{label\_short} & \textit{label\_long} & \textit{material} & \textit{material\_finish} & \textit{description} \\
    \colrule
    \(6785\) & \(8476\) & \(2899\) & \(3375\) & \(11452\) \\
    \botrule
    \end{tabular*}
    \label{tab:dataset_characteristics}
\end{table}

\subsection{Setup}\label{sec:experiments_setup}
We build upon the code base of Dassl \cite{Zhou.2020b,Zhou.2021b} and trained on a single 4090 GPU.
We chose random sampling, an image input size of \(224 \times 224\), and CLIP's pre-trained ViT-B/16 as the image encoder instead of the ResNet version, as ViTs have much less image-specific inductive bias than CNNs. We initialized CoOp with a context length of \(c_{n} = 10\) if not otherwise stated. We trained with a batch size of \(64\) while testing with only \(32\) samples.
This increases the accuracy during validation/testing compared to training accuracy, but we feel this is more realistic in the case of real-world applications contrasting only \(32\) different conceptual object classes in one application. We applied common data augmentation techniques of randomly resizing and cropping as well as flipping edges during training. Besides, we normalized ILID's image data.
We use Adadelta \cite{Zeiler.22.12.2012} with a learning rate of \(0.15\) and a cosine learning rate scheduler with a weight decay of \(1e\text{-}3\). Besides, we used \(3\) warm-up epochs with a constant learning rate of \(1e\text{-}2\) to prevent rapid parameter changes in the initial training stages, which can lead to early overfitting.

\subsection{Quantitative results}\label{sec:experiments_quantitative_results}

\begin{figure*}[!ht]
    \centering
    \includegraphics[width=1.0\linewidth]{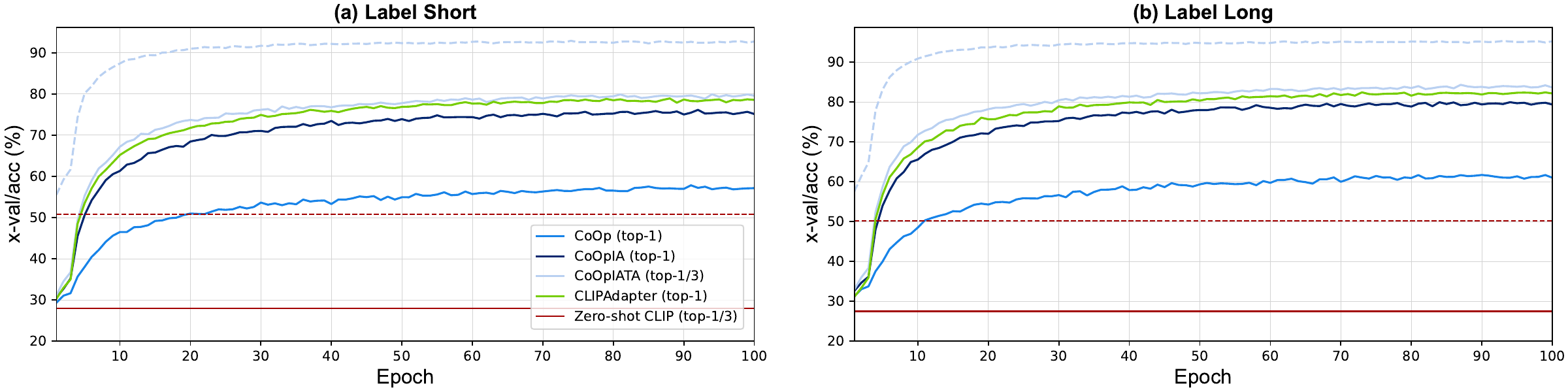}
    \caption{Results of 6-fold cross-validation during transfer learning using different approaches on the ILID.}
    \label{fig:cross_validation}
\end{figure*}

Since we do not have a different real-world language-image dataset at hand, we used 6-fold cross-validation during the evaluation of the different model architectures. Fig. \ref{fig:cross_validation} depicts the validation results of training on the \textit{label\_short} and \textit{label\_long} with CoOp, CoOp + image adapter (CoOpIA) with \(\alpha_i = 0.5\), CoOp + image \(\alpha_i = 0.5\) and text adapter \(\alpha_t = 0.2\) (CoOpIATA), image and text adapter only (CLIPAdapter) (same \(\alpha\)s), and the zero-shot CLIP performance. All accuracies are derived from the top-1 predictions. Additionally, we listed the top-3 accuracies for zero-shot CLIP and CoOpIATA (dashed lines).
~\\

A first observation (\(\triangleright\) Obs. 1) is that all transfer learning approaches effectively outperform CLIP's zero-shot capabilities, even the top-3 accuracies after training for \(\approx 20\) epochs, highlighting that the ILID is out-of-distribution. Even training on the less information-rich \textit{label\_short} outperforms CLIP's zero-shot capabilities. 

CLIP highly depends on the chosen prompt template (\(\triangleright\) Obs. 2). If we look at training on the \textit{label\_long}, the zero-shot (prompt template \textit{"a photo of an industrial product \{label\_long\}"}) accuracy is lower than all other trained methods initialized with random weights in the adapters and CoOp with \textit{"X X X X a photo of an industrial product \{label\_long\}"}, which tokens will be optimized except for the label. CLIP performs poorly if given a prompt that deviates much from the ones during pre-training. That especially accounts for combining the prompt template with multi-word product descriptions.

As expected, the more trainable weights we add, the better the model adapts to the data, while the overall domain generalization to the in-distribution data achieves in the case of \textit{label\_short} and \textit{label\_long} an accuracy of maximum \(79.93 \%\) and \(84.31 \%\), respectively, an image adapter is crucial to effective transfer learning in this case (\(\triangleright\) Obs. 3).

Moreover, the top-performing model also depends on prompt learning, but adapter-styled tuning performs better than only prompt learning on the text stream (\(\triangleright\) Obs. 4). However, adapting to images will reduce the model's performance on slight to out-of-distribution data. That means inference on images that vastly differ from catalog-style ones will definitely have lower performance than the in-distribution images. However, the trained model will still outperform zero-shot CLIP, as we will see in the following sections.

\begin{figure*}[!ht]
    \centering
    \includegraphics[width=1.0\linewidth]{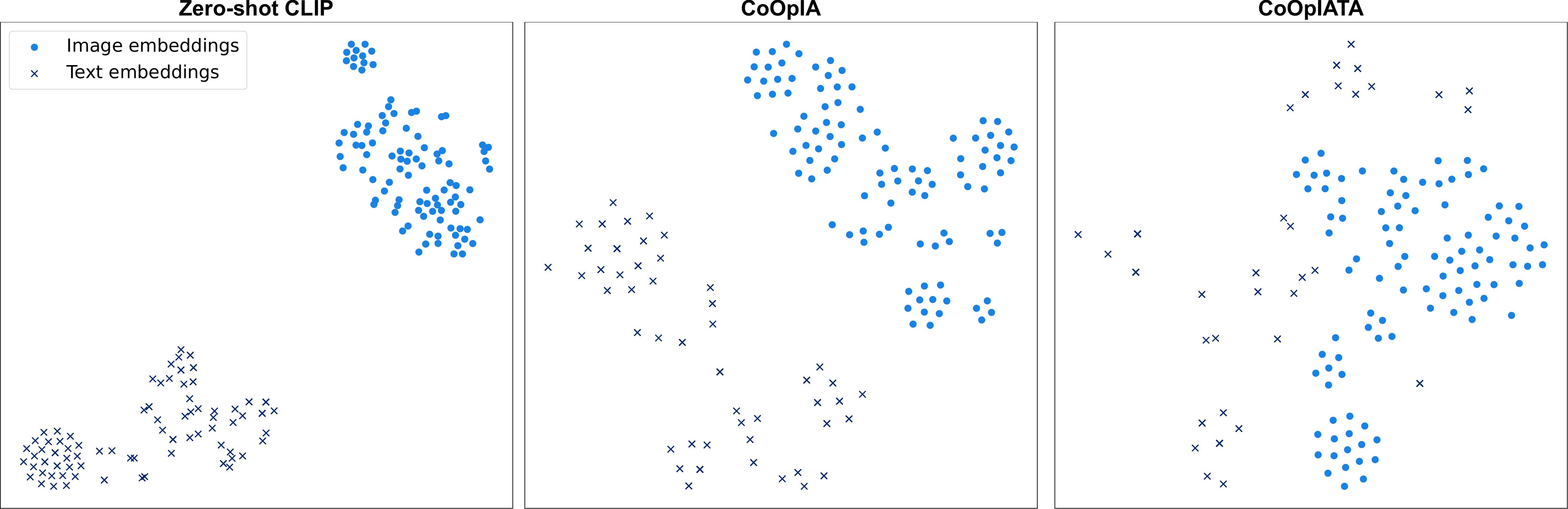}
    \caption{t-SNE diagrams from the same randomly selected $100$ samples (CoOpIA and CoOpIATA were trained for $100$ epochs on the full ILID given the label \textit{label\_short}).}
    \label{fig:tsne_diagram}
\end{figure*}

To gain an understanding of how transfer learning affects the embeddings further, we derived the image and text embeddings after training on the full ILID given the label \textit{label\_short} for \(100\) epochs. Fig. \ref{fig:tsne_diagram} visualizes the high-dimensional embeddings of the same \(100\) samples. With each transfer learning method, adding more trainable weights, the text and image embeddings more jointly share the same embedding space. Further, multiple text embeddings get so close that they are hard to distinguish in the t-SNE diagram at all, which we estimate follows that the transfer learning approaches learn to group semantically close concepts, while in the zero-shot case, these are still more widely clustered.
Moreover, image and text embeddings are much more pronounced after transfer learning than in the zero-shot case.

\subsection{Prompting for materials}\label{sec:material_prompting}
\begin{figure*}[t]
    \centering
    \begin{subfigure}{.19\textwidth}
        \centering
        \includegraphics[width=\linewidth]{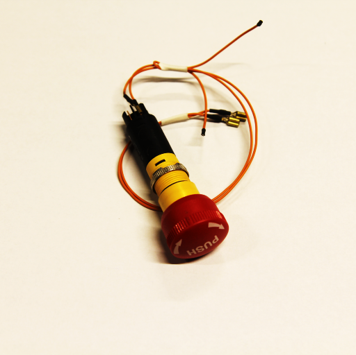}
        \caption{}
        \label{fig:sub1}
    \end{subfigure}%
    \hfill
    \begin{subfigure}{.19\textwidth}
        \centering
        \includegraphics[width=\linewidth]{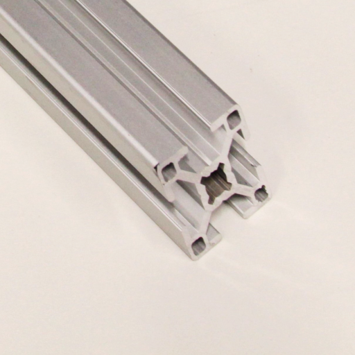}
        \caption{}
        \label{fig:sub2}
    \end{subfigure}
    \hfill
    \begin{subfigure}{.19\textwidth}
        \centering
        \includegraphics[width=\linewidth]{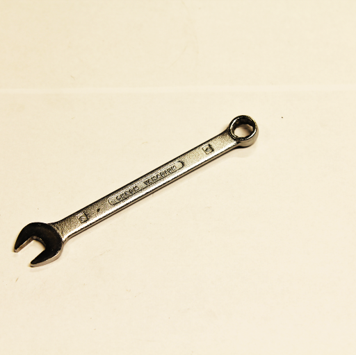}
        \caption{}
        \label{fig:sub3}
    \end{subfigure}
    \hfill
    \begin{subfigure}{.19\textwidth}
        \centering
        \includegraphics[width=\linewidth]{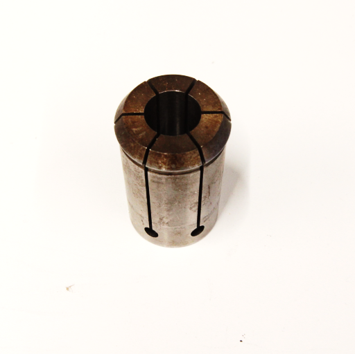}
        \caption{}
        \label{fig:sub4}
    \end{subfigure}%
    \hfill
    \begin{subfigure}{.19\textwidth}
        \centering
        \includegraphics[width=\linewidth]{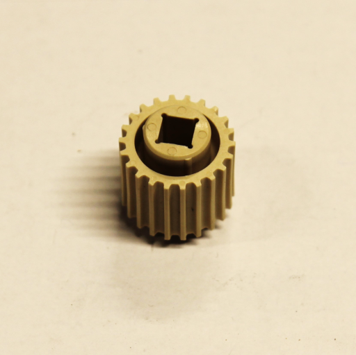}
        \caption{}
        \label{fig:sub5}
    \end{subfigure}
    \caption{Five different real-world images used for prompting material properties.}
    \label{fig:material_prompting}
\end{figure*}

Besides training and testing on the \textit{label\_short} and \textit{label\_long}, we additionally trained CoOpIATA for \(100\) epochs on the \textit{material} label with the initial prompt \textit{"X X X X a photo of an industrial product with material \{\}"}.
We then evaluated the zero-shot and CoOpIATA performance on the images depicted in Fig. \ref{fig:material_prompting} while choosing for the zero-shot test a prompt template similar to for training CoOpIATA.
The results are listed in Tab. \ref{tab:material_prompting}.

Surprisingly, CLIP's zero-shot performance shows \(2\) out of \(5\) true positives, while the transfer learning result in \(5\) out of \(5\).
Further, looking at the scores, we see that our proposed transfer learning method produces much higher confidence in every case, which follows that the different concepts of materials are not in-distribution in the zero-shot case.
Interestingly, a prompt including \textit{\{aluminum\}} results in lower scores than using the word \textit{\{aluminium\}}, which points out that the subtleties or discrepancies of the language used in an industrial context are not mapped after the transfer learning nor in the zero-shot case. That is why we added both words in the prompts.
Further, after transfer learning, judging based on the scores, there is still slight confusion between the material concepts of aluminum and polyamide as well as polyamide and brass. We estimate that the transfer learning introduced a specific object-material-awareness but is still heavily influenced by other image characteristics, like, in our case, the yellow taint.

The given task might not serve a real-world industrial vision use case at this stage, but it shows how ILID can serve different tasks at hand by combining images with different (broad) language information during training. These results again underline a natural language supervised VFM’s rich multimodal capabilities.

\begin{table}[!h]
    \caption{Scores on predicting the object's material properties in the images from Fig. \ref{fig:material_prompting} (bold indicates the highest scores; underlined values correspond to the ground truth).}
    \begin{tabular*}{\hsize}{@{}lccccc@{}}
    \toprule
    & (a) & (b) & (c) & (d) & (e) \\
    \colrule
    
    \multicolumn{6}{l}{Zero-shot CLIP} \\
    \midrule

    \textit{"steel"} & 0.024 & 0.113 & \underline{\textbf{0.330}} & \underline{0.168} & 0.059 \\
    \textit{"polyamide"} & 0.149 & 0.196 & 0.062 & 0.107 & \underline{0.208} \\
    \textit{"thermoplastic"} & 0.245 & 0.141 & 0.050 & 0.034 & 0.097 \\
    \textit{"aluminum or aluminium"} & 0.043 & 0.143 & 0.166 & 0.238 & 0.094 \\
    \textit{"anodized aluminum or} & 0.030 & \underline{0.143} & 0.070 & 0.064 & 0.023 \\
    \textit{aluminium"} & & & & & \\
    \textit{"plastic"} & \underline{\textbf{0.352}} & \textbf{0.244} & 0.099 & 0.107 & \textbf{0.280} \\
    \textit{"brass"} & 0.156 & 0.020 & 0.223 & \textbf{0.282} & 0.240 \\
    
    \midrule
    \multicolumn{6}{l}{CoOpIATA trained on the \textit{material} label} \\
    \midrule

    \textit{"steel"} & 0.007 & 0.033 & \underline{\textbf{0.950}} & \underline{\textbf{0.829}} & 0.137 \\
    \textit{"polyamide"} & 0.135 & 0.368 & 0.004 & 0.008 & \underline{\textbf{0.361}} \\
    \textit{"thermoplastic"} & 0.010 & 0.004 & 0.002 & 0.001 & 0.160 \\
    \textit{"aluminum or aluminium"} & 0.009 & 0.085 & 0.020 & 0.011 & 0.001 \\
    \textit{"anodized aluminum or} & 0.007 & \underline{\textbf{0.374}} & 0.003 & 0.007 & 0.001 \\
    \textit{aluminium"} & & & & & \\
    \textit{"plastic"} & \underline{\textbf{0.694}} & 0.135 & 0.008 & 0.041 & 0.077 \\
    \textit{"brass"} & 0.139 & 0.000 & 0.012 & 0.104 & 0.264 \\
    
    \botrule
    \end{tabular*}
    \label{tab:material_prompting}
\end{table}

\subsection{Language-guided segmentation}\label{sec:experiments_downstream_tasks}

\begin{figure*}[t]
    \centering
    \includegraphics[width=1.0\linewidth]{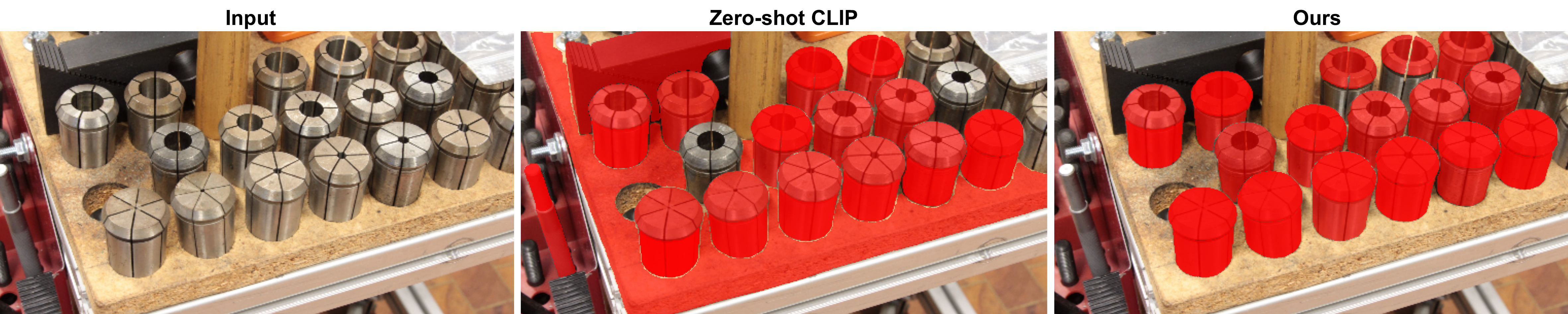}
    \caption{Language-guided segmentation results given prompt \textit{"collet"} compared to zero-shot CLIP under the same settings (segmentation properties and thresholds).}
    \label{fig:segmented_collets}
\end{figure*}

A typical downstream task is a language-guided segmentation utilizing the Segment Anything Model (SAM) \cite{Kirillov.2023}. SAM is a class-agnostic point promptable image segmentation model that outputs hierarchical masks and predicted Intersection over Unions (IoU). Without the need for manual intervention, an automatic mask generation pipeline can sample a point grid and subsequently use Non-Maximum Suppression (NMS) to diminish through merging a large set of masks to form more precise proposals.
In the simplest form, language-guided image segmentation based on SAM and CLIP can be employed by applying CLIP onto all generated masks, which we cut out with a particular delineation factor. CLIP's softmaxed logits can then be thresholded to get the final per-mask class-wise predictions. We only contrasted the object to prompt against an empty class label. Contrasting only two prompts is challenging since the model's overconfidence in one of them is the most pronounced. We chose to do so to avoid any bias by introducing hard negative prompts.

For the language-guided segmentation, we used a CoOpIATA model trained on the complete ILID dataset given the \textit{label\_long} for \(40\) epochs. For completeness, it should be mentioned that we did not compare it against the other approaches, e.g., CLIPAdapter.

Fig. \ref{fig:segmented_collets} depicts the segmentation results in a challenging scene composed of multiple collets stacked on a trolley. The zero-shot results do have many true positives, but overall, we are not able to observe any further prediction patterns. In contrast, the transfer learning approach can effectively distinguish between a mask containing a collet and a mask that does not. 
With only \(17\) word occurrences of \textit{"collet"} in ILID's \textit{label\_long} labels, the resulting model's confidence compared to zero-shot CLIP effectively demonstrates the proposed method. Additionally, the images relating to the labels do not contain collets of the same shapes and sizes, which emphasizes CLIP's learned rich representations.
We discuss two further examples in \ref{app:supplementary_results_segmentation}.

\section{Conclusion and Outlook}\label{sec:conclusion_outlook}

Using VFMs as a building block in an industrial vision application is a promising and transforming technique, improving systems' accuracy, speed, and reliability, e.g., involved in inspection, robotic control, parts identification, and process control, leading to enhanced operational efficiencies and product quality.
As we outlined in Sec. \ref{sec:vfm_industrial_applications}, up to this date, literature only has a limited number of use case ideas regarding using VFMs in industrial applications, which we want to motivate further.

This work strived to make a step towards enabling employing VFM in industrial machine vision applications by introducing the Industrial Language-Image Dataset (ILID) to bring industrial context into CLIP and evaluating effective self-supervised transfer learning from the dataset. We demonstrated this by evaluating downstream tasks from prompting for material properties to language-guided segmentation. With only a limited dataset size of \(\approx 12k\) samples, the results show promising opportunities in machine vision applications when increasing the dataset size or further restricting it to more specific domains.

One can argue that the bigger digital giants like OpenAI or Meta can also incorporate industrial data during the training of their models; however, the overall proposed method from dataset curation to fine-tuning CLIP also suits, e.g., companies with intellectual property constraints or limitations in available computing resources in employing VFMs. Nevertheless, fine-tuning expert models for specific tasks is a common step in creating an AI application, which we, e.g., showcased, given the transfer learning from material properties.
Future work must also elaborate on training with ILID's other labels, like \textit{description}, to further discuss opportunities for other applications.

The current limitations we observed on the text stream are especially the limited learned language subtleties and discrepancies as they occur in industrial contexts. The confusion between the same concept but differently termed in American (aluminium) and British (aluminum) English shows that there is a need for pre-training of the text encoder with broader natural language, e.g., even with extended context, which would enable not only training on shorter image labels.
Further, on the image stream, we observed that the model generalizes well to a variety of an object's different views but does less perform well when contrasting between finer-grained different object types. Here, a custom expert model is probably more suited than transfer learning from a dataset that includes many different object concepts.
The most limiting characteristic is including or inferencing with dimensional quantities, which can hardly be solved when training on images captured with different cameras and their individual intrinsics.

With this work, we hope to encourage the industrial community to employ and work on using VFM in the industrial domain more and more. Therefore, we publicly provide ILID and the code used during training. In the future, we plan to continue increasing the dataset size by incorporating more web catalogs.

\section*{Acknowledgments}
\noindent This work is part of the research project \textit{Intelligent Digital Cabin Twin (InDiCaT)} under the grant number 20D1902C, supported by the \textit{Federal Ministry for Economic Affairs and Climate Action (BMWK)} as part of the \textit{Federal Aeronautical Research Programme LuFo VI-1}.

\noindent We thank \href{https://www.maedler.de/}{MÄDLER GmbH} for granting us the rights to use some of their product images (included in Fig. \ref{fig:overview}, \ref{fig:idea_embedding_space}, \ref{fig:model_architectures}, \ref{fig:excerpt_hinge}, and \ref{fig:excerpt_locking_assembly}) in this publication.

\section*{CRediT author statement}

\noindent K. Moenck: Conceptualization, Methodology, Software, Validation, Formal analysis, Investigation, Resources, Data Curation, Writing – original draft, Writing - review \& editing, Visualization, Supervision, Project administration;
D.T. Thieu: Conceptualization, Methodology, Software, Formal analysis, Investigation, Data Curation, Writing – original draft;
J. Koch: Writing - review \& editing; 
T. Sch{\"u}ppstuhl: Supervision, Project administration, Funding acquisition, Writing - review \& editing.

{
    \small
    \bibliographystyle{elsarticle-num}
    \bibliography{main}
}

\clearpage

\section{Llama-3 prompt}\label{app:llama-prompt}
We followed basic prompt assembly as described for Llama-2 \cite{Touvron.18.07.2023} because up to the date of this publication, there has still been an in-depth explanation of Llama-3 missing. The Llama-2 chat version was trained with a variety of system prompts following patterns like \textit{"You are a helpful, respectful and honest assistant. Always answer as helpfully as possible, while being safe."}, we also included a similar one but tried to include the targeted domain. The brackets \(\{\{\}\}\) point out where we insert the data.

\begin{lstlisting}[basicstyle=\scriptsize,label=lst:llamaprompt_system,caption=System prompt used in the ILID generation pipeline's text transformation step.,float=h,frame=tb,breaklines,breakatwhitespace=true,breakindent=0pt]
You are a helpful assistant for a company that sells industrial products.\n
Do not ask for further details or state additional questions.\n
Do not add additional information or details that are not given by the user.\n
\end{lstlisting}
\begin{lstlisting}[basicstyle=\scriptsize,label=lst:llamaprompt_user,caption=User prompt used in the ILID generation pipeline's text transformation step.,float=h,frame=tb,breaklines,breakatwhitespace=true,breakindent=0pt]
Summarize 'Label: {{}} Text: {{}}'\n
returning the following information: \n
(1) a long label or name of the product without ids, numbers, codes, or sizes 
(2) a short label or name of the product with a maximum of 4 words and shorter than the long label 
(3) description of the product with a maximum of 20 words without ids, numbers, codes, or sizes 
(4) material with a maximum of 5 words
(5) material finish/color with a maximum of 5 words
\end{lstlisting}

\section{Excerpt from the dataset}\label{app:dataset_excerpt}
\begin{figure}[]
    \centering
    \includegraphics[width=1.0\linewidth]{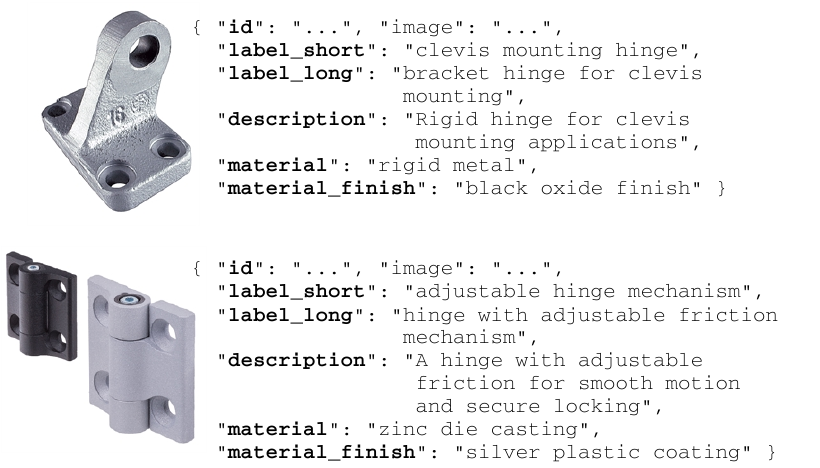}
    \caption{Two samples from the ILID given the keyword \textit{"hinge"}.}
    \label{fig:excerpt_hinge}
\end{figure}

\begin{figure}[]
    \centering
    \includegraphics[width=1.0\linewidth]{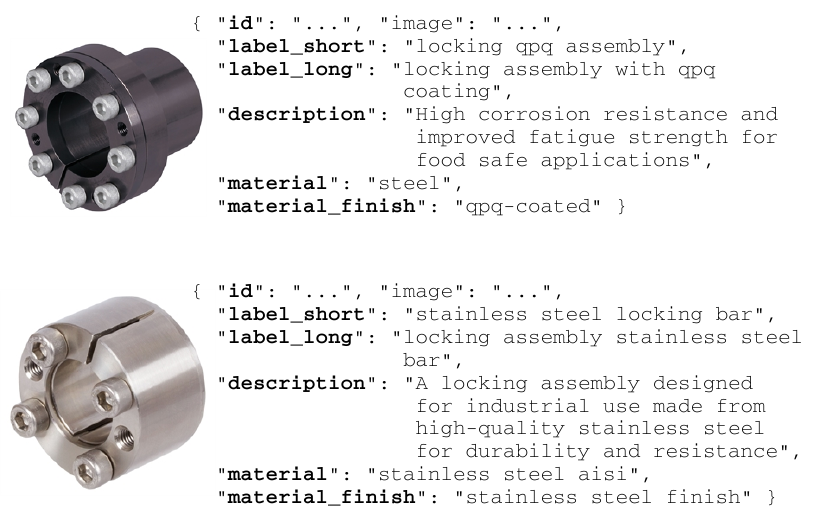}
    \caption{Two samples from the ILID given the keyword \textit{"locking assembly"}.}
    \label{fig:excerpt_locking_assembly}
\end{figure}

Fig. \ref{fig:excerpt_hinge} and Fig. \ref{fig:excerpt_locking_assembly} depict each two samples from the ILID given the keywords \textit{"hinge"} and \textit{"locking assembly"}. Based on the language label, we can observe that the LLM performs differently in extracting the relevant information. As an example, \textit{material} and \textit{material\_finish} confusion occurs when the product page states more than one exact product configuration.

\section{Additional language-guided segmentation results}\label{app:supplementary_results_segmentation}
\begin{figure}[!h]
    \centering
    \includegraphics[width=1.0\linewidth]{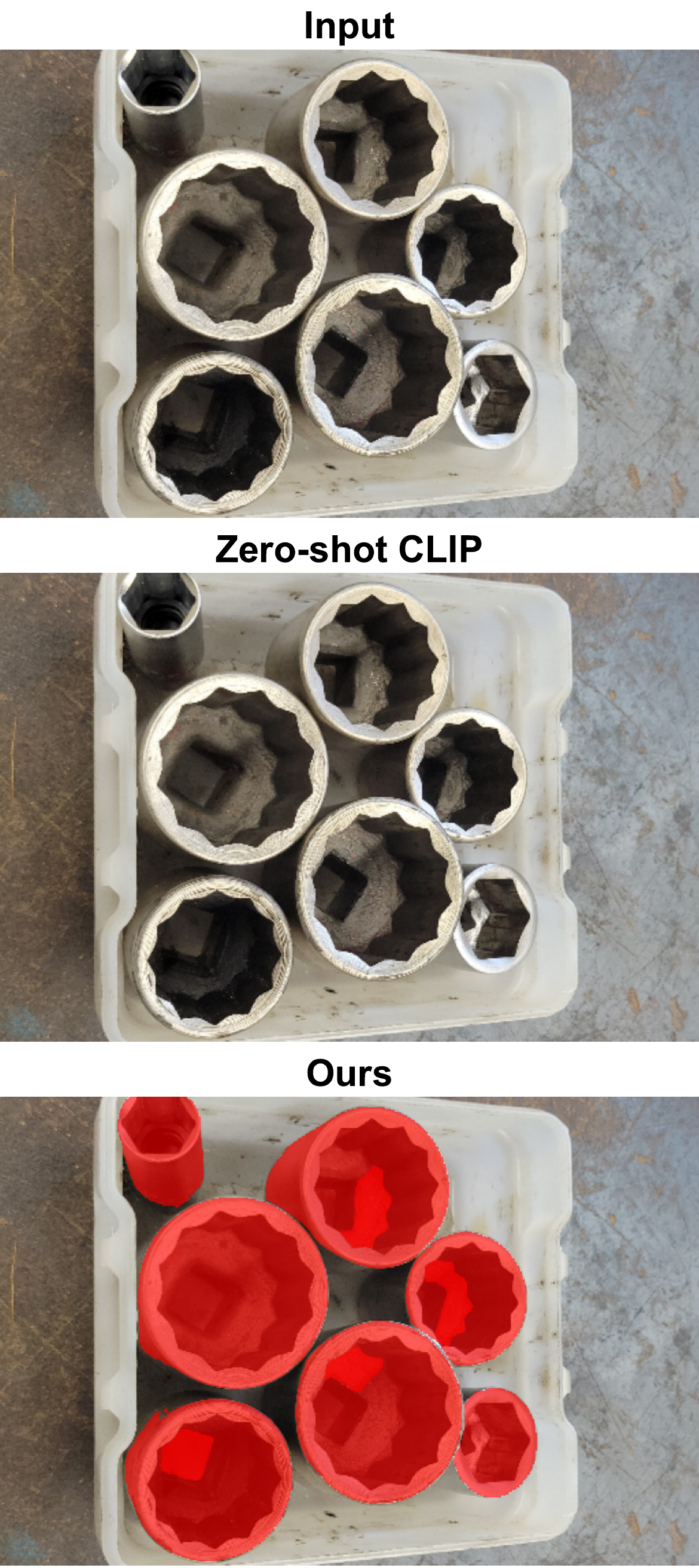}
    \caption{Language-guided segmentation results given the prompt \textit{"socket"} compared to zero-shot CLIP under the same settings.}
    \label{fig:segmented_sockets}
\end{figure}

\begin{figure}[!h]
    \centering
    \includegraphics[width=1.0\linewidth]{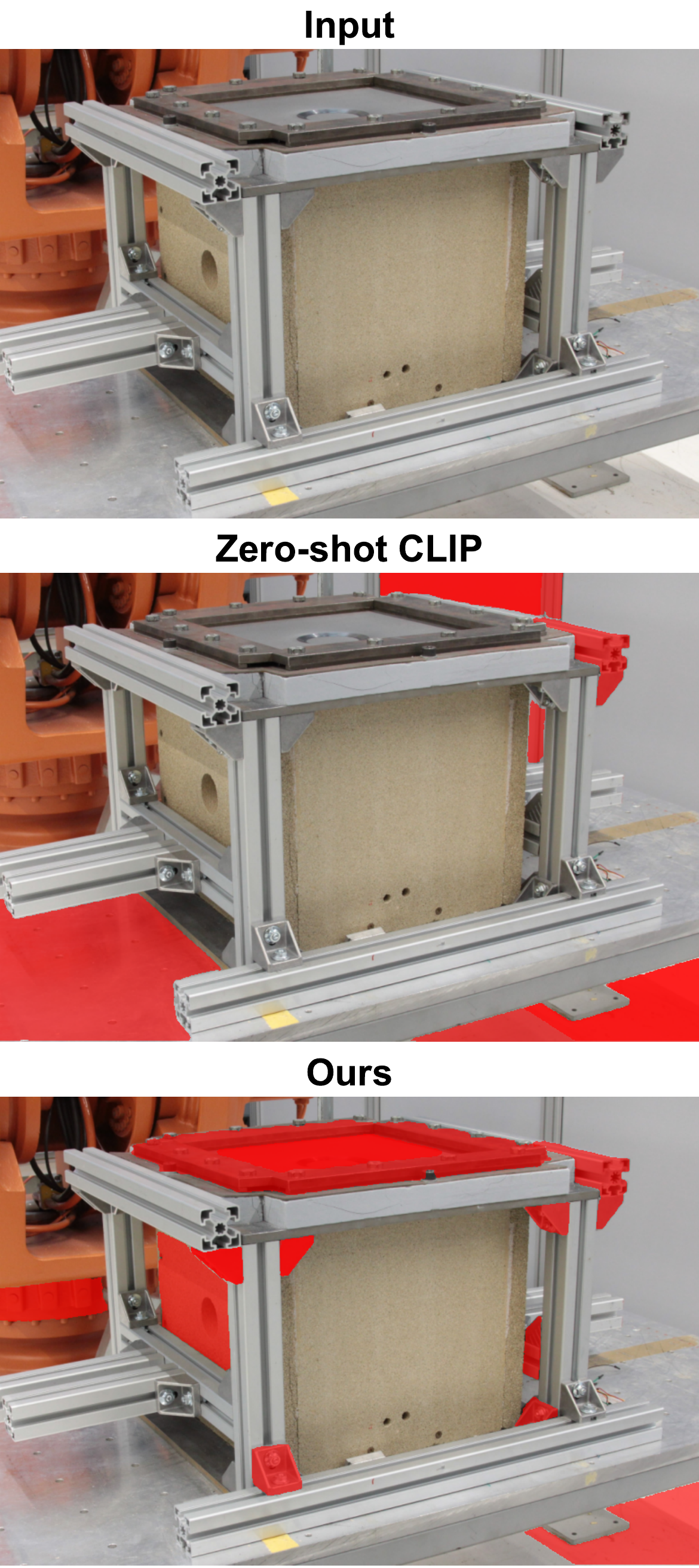}
    \caption{Language-guided segmentation results given the prompt \textit{"bracket for construction profile"} compared to zero-shot CLIP under the same settings.}
    \label{fig:segmented_brackets}
\end{figure}

Fig. \ref{fig:segmented_sockets} and \ref{fig:segmented_brackets} show supplementary results on language-guided image segmentation. In Fig. \ref{fig:segmented_sockets}, we prompted for \textit{"socket"}, whereas zero-shot CLIP does not predict any mask as positive, while our approach segments all sockets.

In Fig. \ref{fig:segmented_brackets}, the results of our most challenging scene are depicted, in which we prompt for \textit{"bracket for construction profile"}. The brackets are imaged far differently than the ones from catalog images, and sometimes they are barely visible. At first sight, the results do not show good performance, especially since we have a few non-detected brackets and a few false positive predictions. We explain the false positive on the top with the cropping strategy, while we have no explanation for the false predictions on the lower right.
The false positives can result from the axis-preserving cropping strategy of the used method, in which a cropped segment includes parts of the surroundings. A lot of the false positive segments contain parts of brackets. Employing more sophisticated language-image segmentation methods, like \cite{Wang.23.10.2023}, based on CLIP and SAM that do not rely on such a straightforward cropping strategy could prevent such wrongful predictions.
In contrast, we observed less performance during the segment classification with CLIP when the background was not included in the segments.

\end{document}